\documentclass[10pt,twocolumn]{article}

\usepackage[utf8]{inputenc}
\usepackage[T1]{fontenc}
\usepackage[margin=0.75in]{geometry}
\usepackage{graphicx}
\usepackage{amsmath,amssymb}
\usepackage{booktabs}
\usepackage{tabularx}
\usepackage{hyperref}
\usepackage{url}
\usepackage{xcolor}
\usepackage{caption}
\usepackage{subcaption}
\usepackage{float}
\usepackage{tikz}
\usetikzlibrary{shapes.geometric, arrows.meta, positioning, fit, backgrounds}
\usepackage{algorithm}
\usepackage{algorithmic}
\usepackage[numbers,sort&compress]{natbib}
\usepackage{authblk}
\usepackage{titlesec}
\usepackage{enumitem}
\usepackage{multirow}

\hypersetup{
    colorlinks=true,
    linkcolor=blue!70!black,
    citecolor=blue!70!black,
    urlcolor=blue!70!black
}

\titlespacing*{\section}{0pt}{1.2em}{0.5em}
\titlespacing*{\subsection}{0pt}{0.8em}{0.3em}
\titlespacing*{\subsubsection}{0pt}{0.6em}{0.2em}
\setlist{nosep, leftmargin=1.2em}

\title{\Large \textbf{MemMachine: A Ground-Truth-Preserving Memory System\\for Personalized AI Agents}}

\author[1]{Shu Wang\thanks{shu.wang@memverge.com}}
\author[1]{Edwin Yu\thanks{edwin.yu@memverge.com}}
\author[1]{Oscar Love\thanks{oscar.love@memverge.com}}
\author[1]{Tom Zhang\thanks{tom.zhang@memverge.com}}
\author[1]{Tom Wong\thanks{tom.wong@memverge.com}}
\author[1]{Steve Scargall\thanks{steve.scargall@memverge.com}}
\author[1]{Charles Fan\thanks{charles.fan@memverge.com}}
\affil[1]{MemVerge, Inc.}

\date{March 2026}

\begin{document}
\maketitle

\begin{abstract}
Large language model (LLM) agents require persistent memory to maintain personalization, factual continuity, and long-horizon task performance, yet standard context-window and retrieval-augmented generation (RAG) workflows remain brittle under multi-session interactions. We present MemMachine, an open-source memory system that combines short-term memory, long-term episodic memory, and profile memory in a ground-truth-preserving architecture that stores raw conversational episodes and minimizes routine LLM-based extraction. MemMachine introduces contextualized retrieval that expands nucleus matches with neighboring episode context, improving recall for conversational queries where semantically related evidence is distributed across turns. 

Across multiple benchmarks, MemMachine demonstrates strong accuracy-efficiency tradeoffs. On LoCoMo, it achieves an overall score of 0.9169 with gpt-4.1-mini. On LongMemEval\textsubscript{S} (ICLR~2025), a systematic ablation study across six optimization dimensions yields \textbf{93.0\%} overall accuracy, with retrieval-stage optimizations contributing more than ingestion-stage changes: retrieval depth tuning (+4.2\%), context formatting (+2.0\%), search prompt design (+1.8\%), and query bias correction (+1.4\%) each outweigh sentence chunking (+0.8\%). A surprising finding is that GPT-5-mini outperforms GPT-5 as the answer LLM (+2.6\%) when paired with optimized prompts, yielding the most cost-efficient configuration. In matched memory-mode comparisons, MemMachine uses approximately 80\% fewer input tokens than Mem0. We further evaluate a Retrieval Agent that routes queries to direct retrieval, parallel decomposition, or iterative chain-of-query strategies, reaching 93.2\% on HotpotQA hard and 92.6\% on WikiMultiHop under randomized-noise conditions. 

These results suggest that preserving episodic ground truth while layering adaptive retrieval strategies yields robust long-term memory behavior for personalized agents.
\end{abstract}

\begin{table}[t]
\centering
\caption{Benchmark snapshot (quick overview). Detailed setup and full breakdowns are provided in later sections.}
\label{tab:first_page_snapshot}
\scriptsize
\setlength{\tabcolsep}{5pt}
\begin{tabular}{@{}lcc@{}}
\toprule
\textbf{Benchmark} & \textbf{Metric} & \textbf{Result} \\
\midrule
LoCoMo & Overall score (gpt-4.1-mini) & \textbf{91.69\%} \\
LongMemEval\textsubscript{S} & Ablation best (gpt-5-mini) & \textbf{93.0\%} \\
HotpotQA hard & Retrieval Agent accuracy & \textbf{93.2\%} \\
WikiMultiHop$^\dagger$ & Retrieval Agent accuracy & \textbf{92.6\%} \\
Mem0 comparison & Input token reduction & \textbf{\textasciitilde80\% less} \\
\bottomrule
\multicolumn{3}{@{}l@{}}{\scriptsize $^\dagger$Randomized-noise setting.} \\
\end{tabular}
\end{table}

\section{Introduction}
\label{sec:intro}

Transformer-based large language models (LLMs) have become the computational foundation for a rapidly growing class of autonomous AI applications, from conversational assistants and customer-facing agents to complex multi-agent workflows~\cite{packer2024memgpt, park2023generative}. Despite their broad capabilities, LLMs exhibit two fundamental limitations that constrain their utility in persistent, personalized applications:

\begin{enumerate}
    \item \textbf{Static Parameters.} Once trained, an LLM's weights are fixed. The model cannot acquire new knowledge from interactions without costly fine-tuning or re-training.
    \item \textbf{Restricted Context Window.} LLMs operate within a finite context window, requiring applications to carefully curate and compress inference data, often at the cost of losing relevant historical context.
\end{enumerate}

Retrieval-Augmented Generation (RAG)~\cite{lewis2020rag} has emerged as the dominant paradigm for injecting external knowledge into LLM workflows. However, conventional RAG architectures are designed for static document collections and do not support the dynamic, bidirectional interactions characteristic of AI agents that must learn from, and adapt to, evolving user contexts across sessions.

What AI agents require is a \emph{memory system}—a mechanism that goes beyond document retrieval to store, organize, recall, and reason over past experiences. Drawing on established models from cognitive science~\cite{tulving1972episodic, atkinson1968human}, such a system should provide:

\begin{itemize}
    \item \textbf{Short-term memory (STM):} An immediate workspace maintaining current context, with limited capacity.
    \item \textbf{Long-term episodic memory:} A store of specific past experiences, providing ground truth about what occurred.
    \item \textbf{Semantic memory:} High-level summaries, facts, and user profiles distilled from raw experience.
    \item \textbf{Procedural memory:} Learned rules, strategies, and action patterns that guide agent behavior.
\end{itemize}

A growing body of systems have begun to address this challenge. MemGPT~\cite{packer2024memgpt} explored virtual memory management for LLMs. Mem0~\cite{chhikara2025mem0} and Zep~\cite{rasmussen2025zep} provide long-term memory layers for AI agents. However, these systems primarily rely on LLMs for data extraction, update, aggregation, and deletion—a design that introduces high operational cost, accuracy concerns from probabilistic extraction, and compounding error over time.

\subsection{Contributions}

In this paper, we present \textbf{MemMachine}, an open-source memory system that takes a fundamentally different approach. Our key contributions are:

\begin{enumerate}
    \item \textbf{Ground-truth-preserving architecture.} MemMachine stores raw conversational episodes and indexes them at the sentence level, minimizing LLM dependence for routine memory operations and preserving factual integrity.
    \item \textbf{Contextualized retrieval.} A novel retrieval mechanism that expands nucleus episodes with neighboring context to form episode clusters, addressing the embedding dissimilarity problem inherent in conversational data.
    \item \textbf{Cost-efficient operation.} By reserving LLM calls for summarization and high-level abstraction rather than per-message extraction, MemMachine achieves approximately 80\% reduction in token usage compared to competing systems.
    \item \textbf{Personalization support.} A profile memory system that extracts and maintains user preferences, facts, and behavioral patterns to enable personalized agent interactions across sessions.
    \item \textbf{Leading LoCoMo performance (as of March 2026).} MemMachine achieves 0.9169 on LoCoMo with gpt-4.1-mini, among the strongest published results for open memory frameworks and above reported Mem0, Zep, Memobase, LangMem, and OpenAI baseline scores.
    \item \textbf{LongMemEval\textsubscript{S} ablation study.} A systematic evaluation on LongMemEval (ICLR~2025) across six optimization dimensions---sentence chunking, query bias correction, context formatting, retrieval depth, search prompt design, and answer-model selection---achieving 93.0\% overall accuracy and revealing that retrieval-stage optimizations dominate over ingestion-stage changes.
    \item \textbf{Comprehensive evaluation.} We provide reproducible benchmark scripts and analyze the impact of embedding models, reranking strategies, LLM model selection, and retrieval parameters on memory performance.
    \item \textbf{Retrieval Agent for multi-hop reasoning.} An LLM-orchestrated retrieval pipeline that classifies queries into structural types and routes them to purpose-built strategies (direct search, parallel decomposition, or iterative chain-of-query), achieving 93.2\% accuracy on the HotpotQA hard set while maintaining bounded cost.
\end{enumerate}

MemMachine is released under the Apache 2.0 license and is available at \url{https://github.com/MemMachine/MemMachine}.

\section{Related Work}
\label{sec:related}

\subsection{Memory for AI Agents}

The need for persistent memory in LLM-based agents has been recognized across multiple research threads. Hu et al.~\cite{hu2025memory_survey} provide a comprehensive survey organizing agent memory by forms (token-level, parametric, latent), functions (factual, experiential, working), and dynamics (formation, evolution, retrieval). Park et al.~\cite{park2023generative} demonstrated the power of memory in generative agents that simulate human behavior, using a memory stream architecture with retrieval, reflection, and planning.

MemGPT~\cite{packer2024memgpt} introduced an operating-system-inspired virtual memory hierarchy for LLMs, managing context by paging information between a main context and external storage. While pioneering, MemGPT's approach requires complex memory management that can introduce latency and depends on LLM-driven decisions for memory operations.

\subsection{Existing Memory Systems}

\textbf{Mem0}~\cite{chhikara2025mem0} provides a production-oriented memory layer that extracts facts from conversations using LLM calls, stores them in hybrid vector and graph databases, and retrieves them for agent inference. While effective, the per-message LLM extraction approach incurs significant cost and can introduce factual drift through accumulated extraction errors.

\textbf{Zep}~\cite{rasmussen2025zep} implements a temporal knowledge graph architecture that tracks how facts evolve over time, combining graph-based memory with vector search. Zep excels at relationship modeling and temporal reasoning but introduces complexity in deployment and configuration.

\textbf{Memobase}\footnote{\url{https://github.com/memodb-io/memobase}} and \textbf{LangMem}\footnote{\url{https://github.com/langchain-ai/langmem}} represent additional approaches, with Memobase providing structured memory storage and LangMem integrating memory into the LangChain ecosystem.

More recently, \textbf{Mastra}~\cite{mastra2026observational} introduced \emph{observational memory}, which uses two background agents (Observer and Reflector) to compress conversation history into a dated observation log that stays in context, eliminating retrieval entirely. This approach achieves strong LongMemEval scores (94.87\% with GPT-5-mini) and enables aggressive prompt caching, but trades away the ability to search a broader external corpus—making it less suitable for open-ended knowledge discovery or compliance-heavy recall use cases where ground truth access is required.

\textbf{MemOS}~\cite{li2025memos} takes the most ambitious architectural stance, proposing a full \emph{memory operating system} that treats memory as a first-class schedulable resource analogous to CPU or storage in traditional operating systems. MemOS unifies three memory types under a single abstraction called \emph{MemCube}: \emph{plaintext memory} (externally injected text, akin to RAG), \emph{activation memory} (KV-cache states from inference), and \emph{parametric memory} (knowledge embedded in model weights, e.g., LoRA adapters). MemCubes carry rich metadata including provenance, versioning, access policies, and lifecycle state, enabling cross-type transformations—for example, promoting frequently accessed plaintext into KV-cache templates for faster inference, or distilling stable knowledge into parametric weights. The system implements a three-layer architecture (interface, operation, infrastructure) with a memory scheduler that orchestrates predictive preloading and multi-user session management. On LoCoMo, MemOS reports 75.80 using GPT-4o-mini as the processing LLM, with a claimed 159\% improvement in temporal reasoning over OpenAI's global memory.

While MemOS represents a compelling long-term vision—particularly its memory lifecycle governance and cross-type transformation pathways—its scope is significantly broader than the other systems discussed here. The parametric and activation memory types require direct access to model internals (weights, KV-caches), which limits portability across LLM providers and closed-source APIs. By contrast, MemMachine, Mem0, Zep, and Mastra operate at the \emph{application layer}, interfacing with LLMs through standard text-based APIs, which enables them to work with any model provider without modification.

These diverse approaches highlight a fundamental design tension in agent memory: \emph{compression vs.\ preservation}. Systems that aggressively compress (Mastra, Mem0) achieve smaller context windows and lower per-query cost but risk losing critical detail. Systems that preserve raw data (MemMachine) maintain factual integrity at the cost of requiring efficient retrieval mechanisms. A related tension exists between \emph{retrieval-based} approaches (MemMachine, Mem0, Zep) that search selectively and \emph{context-based} approaches (Mastra) that keep compressed history always in context. MemOS introduces a third dimension: \emph{memory-layer depth}, spanning from application-level text memory through inference-level activation caching to model-level parametric adaptation. Table~\ref{tab:design_space} compares these architectural choices.

\subsection{Memory Benchmarks}

The evaluation landscape for agent memory has matured significantly. \textbf{LoCoMo}~\cite{maharana2024locomo} evaluates very long-term conversational memory through multi-session dialogues with single-hop, multi-hop, temporal, and open-domain question types. \textbf{LongMemEval}~\cite{wu2024longmemeval} benchmarks five core long-term memory abilities—information extraction, multi-session reasoning, temporal reasoning, knowledge updates, and abstention—using scalable chat histories. \textbf{EpBench}~\cite{huet2025epbench} focuses on episodic memory evaluation through synthetic narrative corpora ranging from 100K to 1M tokens.


\subsection{Memory Types in Cognitive Science}

Our design draws on the established taxonomy from cognitive science. Tulving~\cite{tulving1972episodic} distinguished \emph{episodic memory} (memory of specific personal experiences bound to time and place) from \emph{semantic memory} (general knowledge abstracted from experience). The Atkinson-Shiffrin model~\cite{atkinson1968human} formalized the multi-store architecture with sensory, short-term, and long-term stores. In the AI agent context, we adopt these distinctions while acknowledging that current implementations approximate rather than replicate human memory processes.

\section{Memory Types for AI Agents}
\label{sec:memory_types}

AI agent memory systems draw inspiration from cognitive science while adapting to the practical requirements of LLM-based applications. This section describes the primary memory types and their roles, with emphasis on those implemented in MemMachine.

\subsection{Episodic Memory}

Episodic memory stores specific past experiences—\emph{what} happened, \emph{when}, \emph{where}, and with \emph{whom}. In the agent context, each conversational turn or interaction constitutes an \emph{episode}, a discrete unit of experience with associated metadata (timestamp, participants, session identifier).

Episodic memory serves as ground truth. When an agent needs to recall what a user said, what was decided, or what sequence of events occurred, it queries episodic memory for the raw record. This is essential for factual accuracy, auditability, and trust.

\textbf{When to use:} Factual recall, reconstructing conversation history, answering questions about specific past interactions, providing evidence for decisions, and maintaining conversational continuity across sessions.

\subsection{Semantic Memory (Profile Memory)}

Semantic memory stores generalized knowledge abstracted from episodic experience—user preferences, facts, behavioral patterns, and stable attributes. In MemMachine, this is implemented as \textbf{Profile Memory}, which extracts and maintains structured user profiles from conversational data.

Unlike episodic memory, which preserves the raw record, semantic/profile memory distills high-level patterns: ``The user prefers vegetarian restaurants,'' ``The user works in financial services,'' or ``The user's preferred communication style is concise and technical.''

\textbf{When to use:} Personalization, preference-aware recommendations, adapting tone and content, and providing context that does not require recalling a specific episode.

\subsection{Procedural Memory}

Procedural memory encodes learned skills, strategies, and behavioral rules—\emph{how} to do things. In agent systems, this includes tool-use patterns, workflow steps, and decision heuristics. MemMachine does not currently implement procedural memory, though the architecture can be extended to support it.

\textbf{When to use:} Multi-step task execution, tool selection, workflow automation, and strategy reuse.

\subsection{Temporal Awareness}

While not a separate memory type, temporal awareness is a cross-cutting capability. MemMachine tags all episodes with timestamps and supports temporal filtering during search, enabling agents to reason about event ordering, recency, and duration. This provides limited but valuable temporal reasoning without requiring a dedicated temporal memory module.

\subsection{Episodic vs.\ Semantic: When to Use Each}

The choice between episodic and semantic retrieval depends on the query type:

\begin{table}[H]
\centering
\small
\caption{Episodic vs.\ Semantic Memory Usage}
\label{tab:ep_vs_sem}
\begin{tabular}{@{}p{2.1cm}p{2.5cm}p{2.5cm}@{}}
\toprule
\textbf{Criterion} & \textbf{Episodic} & \textbf{Semantic} \\
\midrule
Query type & Specific past events & General preferences \\
Accuracy need & Ground truth & Approximate \\
Temporal scope & Point-in-time & Cross-session \\
Data form & Raw conversation & Extracted facts \\
Example & ``What did I say about X?'' & ``What foods do I like?'' \\
\bottomrule
\end{tabular}
\end{table}

In practice, effective agents combine both: episodic memory for factual grounding and semantic/profile memory for personalization.

\section{MemMachine Architecture}
\label{sec:architecture}

MemMachine implements a client-server architecture with a two-tier memory system comprising episodic memory (short-term and long-term) and profile memory (semantic). This section describes the system's components, data flow, and key design decisions.

\subsection{System Overview}

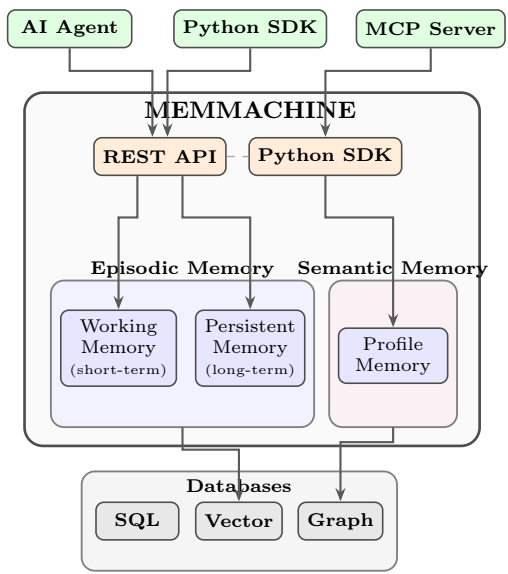
\begin{figure}[t]
\centering
\begin{tikzpicture}[
    every node/.style={font=\scriptsize},
    clientbox/.style={rectangle, draw=black!70, rounded corners=3pt, minimum width=1.65cm, minimum height=0.5cm, font=\scriptsize\bfseries, align=center, fill=green!12, line width=0.6pt},
    apibox/.style={rectangle, draw=black!70, rounded corners=3pt, minimum width=1.65cm, minimum height=0.5cm, font=\scriptsize\bfseries, align=center, fill=orange!15, line width=0.6pt},
    membox/.style={rectangle, draw=black!70, rounded corners=3pt, minimum width=1.45cm, minimum height=0.7cm, font=\scriptsize, align=center, fill=blue!10, line width=0.6pt},
    groupbox/.style={rectangle, draw=black!50, rounded corners=5pt, line width=0.7pt},
    dbbox/.style={rectangle, draw=black!70, rounded corners=3pt, minimum width=1.1cm, minimum height=0.5cm, font=\scriptsize\bfseries, align=center, fill=gray!18, line width=0.6pt},
    arrow/.style={-{Stealth[length=1.8mm]}, thick, black!65},
]

\coordinate (mm-tl) at (-0.6, 0.35);
\coordinate (mm-br) at (5.45, -4.35);
\fill[gray!4, rounded corners=7pt] (mm-tl) rectangle (mm-br);
\draw[black!70, rounded corners=7pt, line width=1.0pt] (mm-tl) rectangle (mm-br);
\node[font=\small\bfseries] at (2.4, 0.12) {MEMMACHINE};

\fill[blue!5, rounded corners=5pt] (-0.25, -2.15) rectangle (3.25, -4.1);
\draw[black!50, rounded corners=5pt, line width=0.7pt] (-0.25, -2.15) rectangle (3.25, -4.1);
\node[font=\scriptsize\bfseries] at (1.5, -2.0) {Episodic Memory};

\fill[purple!6, rounded corners=5pt] (3.45, -2.15) rectangle (5.15, -4.1);
\draw[black!50, rounded corners=5pt, line width=0.7pt] (3.45, -2.15) rectangle (5.15, -4.1);
\node[font=\scriptsize\bfseries] at (4.3, -2.0) {Semantic Memory};

\node[clientbox] (agent) at (0.0, 1.2) {AI Agent};
\node[clientbox] (pysdk) at (2.4, 1.2) {Python SDK};
\node[clientbox] (mcp) at (4.8, 1.2) {MCP Server};

\node[apibox] (restapi) at (1.2, -0.5) {REST API};
\node[apibox] (sdkapi) at (3.4, -0.5) {Python SDK};
\draw[dashed, black!35, line width=0.5pt] (restapi.east) -- (sdkapi.west);

\node[membox] (wm) at (0.65, -3.05) {Working\\Memory\\{\tiny(short-term)}};
\node[membox] (persistent) at (2.4, -3.05) {Persistent\\Memory\\{\tiny(long-term)}};
\node[membox] (profile) at (4.3, -3.15) {Profile\\Memory};

\node[dbbox] (sql) at (0.9, -5.35) {SQL};
\node[dbbox] (vector) at (2.25, -5.35) {Vector};
\node[dbbox] (graph) at (3.6, -5.35) {Graph};

\begin{scope}[on background layer]
\node[groupbox, fill=gray!8, fit=(sql)(vector)(graph), inner xsep=0.18cm, inner ysep=0.4cm] (dbgroup) {};
\end{scope}
\node[font=\scriptsize\bfseries] at (2.25, -4.88) {Databases};

\draw[arrow] (agent.south) -- ++(0,-0.3) -| ([xshift=-0.1cm]restapi.north);
\draw[arrow] (pysdk.south) -- ++(0,-0.3) -| ([xshift=0.1cm]restapi.north);
\draw[arrow] (mcp.south) -- ++(0,-0.3) -| (sdkapi.north);

\draw[arrow] ([xshift=-0.3cm]restapi.south) -- ++(0,-0.55) -| (wm.north);
\draw[arrow] ([xshift=0.3cm]restapi.south) -- ++(0,-0.55) -| (persistent.north);
\draw[arrow] (sdkapi.south) -- ++(0,-0.55) -| (profile.north);

\draw[arrow] (1.5, -4.1) -- ++(0,-0.3) -| (vector.north);
\draw[arrow] (4.3, -4.1) -- ++(0,-0.2) -| (graph.north);

\end{tikzpicture}
\caption{MemMachine system architecture. Clients (AI agents, Python SDK, MCP servers) access MemMachine through REST API and SDK interfaces. Internally, episodic memory comprises working memory (short-term) and persistent memory (long-term), while semantic memory stores user profiles. All memory types are backed by SQL, vector, and graph databases.}
\label{fig:architecture}
\end{figure}

Figure~\ref{fig:architecture} illustrates the high-level architecture. Agents interact with MemMachine through three API interfaces: a RESTful API (v2), a Python SDK, and a Model Context Protocol (MCP) server. The server manages memory processing, while the storage layer persists data across PostgreSQL (with pgvector for vector search), SQLite, and Neo4j (for graph-structured long-term memory).

\subsection{Data Ingestion}

Raw messages are submitted to MemMachine with metadata. The system organizes each message into an internal data structure called an \textbf{Episode}. Each episode represents one conversational turn and carries required metadata:

\begin{itemize}
    \item \textbf{Producer:} The source of the message (user, agent, system).
    \item \textbf{Timestamp:} When the message was produced.
    \item \textbf{Session ID:} Grouping episodes into conversational sessions.
    \item \textbf{Custom metadata:} Arbitrary key-value pairs for domain-specific filtering.
\end{itemize}

Episodes are stored in a central database as a raw data repository and simultaneously dispatched to episodic memory and profile memory for ingestion and indexing.

\subsection{Short-Term Memory}

Short-term memory (STM) maintains a configurable context window of the most recent episodes, providing immediate conversational context. Key behaviors:

\begin{itemize}
    \item Holds a predefined number of recent episodes.
    \item Generates compressed summaries of session-level interactions via LLM-based abstraction.
    \item When content exceeds the window, both episodes and summaries are compressed for efficient storage and eventually transferred to long-term memory.
\end{itemize}

STM ensures that agents always have access to the immediate conversational context without requiring a retrieval step, while the summarization mechanism preserves the gist of older context within the window.

\subsection{Long-Term Memory}

Long-term memory (LTM) provides persistent, searchable storage for all episodes that have exited the STM window. The indexing pipeline comprises four stages:

\begin{enumerate}
    \item \textbf{Sentence Extraction:} Each episode is segmented into individual sentences using NLTK's Punkt tokenizer~\cite{kiss2006unsupervised}. This fine-grained decomposition enables precise embedding and retrieval at the sentence level rather than the episode level.
    \item \textbf{Metadata Augmentation:} Each sentence inherits metadata from its parent episode (timestamp, producer, session) and receives a unique identifier.
    \item \textbf{Relational Mapping:} Sentences are linked to their originating episodes, maintaining provenance.
    \item \textbf{Embedding Generation:} Semantic embeddings are generated for each sentence. MemMachine supports configurable embedding models, enabling domain-specific models for improved performance.
\end{enumerate}

The original episodes, augmented sentences, and embeddings are persisted in the database. Neo4j provides graph-based storage that enables relational traversal, while PostgreSQL with pgvector supports efficient vector similarity search.

\subsection{Memory Search and Recall}

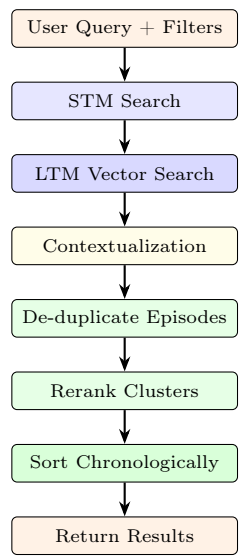
\begin{figure}[t]
\centering
\begin{tikzpicture}[
    node distance=0.45cm,
    box/.style={rectangle, draw, rounded corners=2pt, minimum width=3cm, minimum height=0.5cm, font=\scriptsize, align=center, fill=#1},
    box/.default={blue!8},
    arrow/.style={-{Stealth[length=2mm]}, thick}
]

\node[box=orange!10] (query) {User Query + Filters};
\node[box=blue!10, below=of query] (stm_search) {STM Search};
\node[box=blue!15, below=of stm_search] (ltm_search) {LTM Vector Search};
\node[box=yellow!10, below=of ltm_search] (context) {Contextualization};
\node[box=green!10, below=of context] (dedup) {De-duplicate Episodes};
\node[box=green!10, below=of dedup] (rerank) {Rerank Clusters};
\node[box=green!15, below=of rerank] (sort) {Sort Chronologically};
\node[box=orange!10, below=of sort] (result) {Return Results};

\draw[arrow] (query) -- (stm_search);
\draw[arrow] (stm_search) -- (ltm_search);
\draw[arrow] (ltm_search) -- (context);
\draw[arrow] (context) -- (dedup);
\draw[arrow] (dedup) -- (rerank);
\draw[arrow] (rerank) -- (sort);
\draw[arrow] (sort) -- (result);

\end{tikzpicture}
\caption{Memory recall workflow. The query passes through STM, LTM vector search, contextualization, deduplication, reranking, and chronological sorting before returning results.}
\label{fig:search_flow}
\end{figure}

Memory search in MemMachine follows a staged recall pipeline that balances speed, coverage, and factual grounding. The system first checks near-term context, then expands into long-term episodic retrieval, and finally refines candidates before returning results. Figure~\ref{fig:search_flow} (Figure 2) summarizes this end-to-end workflow.

\subsubsection{Long-Term Memory Search}

LTM search begins with a vector similarity search using Approximate Nearest Neighbor (ANN) or exact matching against the sentence embeddings. Matched sentences are traced back to their originating episodes, with duplicates removed. The system then applies \emph{contextualization} (Section~\ref{sec:contextualization}).

\subsubsection{Episodic Memory Search}

Episodic search coordinates STM and LTM. It invokes LTM search for episodes outside the STM window, deduplicates overlapping episodes between STM and LTM, sorts all results chronologically to preserve conversational flow, and returns STM episodes, the STM summary, and LTM episodes for agent-driven context assembly.

\subsection{Contextualization}
\label{sec:contextualization}

A key challenge in conversational memory retrieval is that contextually important episodes may have embeddings quite dissimilar from the query. Unlike document retrieval in traditional RAG, where each chunk is relatively self-contained, conversational turns are highly interdependent. A question about ``the restaurant recommendation'' requires not just the turn containing the recommendation, but the surrounding turns that establish what was asked, why, and what constraints were given.

MemMachine addresses this with \textbf{contextualized retrieval}:

\begin{enumerate}
    \item The \textbf{nucleus episode} is located via embedding search.
    \item Immediate \textbf{neighboring episodes} are retrieved (one preceding, two following) to form an \textbf{episode cluster}.
    \item Episode clusters are \textbf{reranked} using a cross-encoder or other reranking model.
    \item The top-$k$ clusters are provided for LLM inference.
\end{enumerate}

This approach ensures that the LLM receives not just the most semantically similar turns, but the conversational context necessary for accurate reasoning. Our experimental results (Section~\ref{sec:results}) demonstrate the significant impact of contextualization on benchmark performance.

\subsection{Profile Memory (Semantic Memory)}

Profile memory extracts and maintains user-level facts and preferences from conversational data. Unlike episodic memory, which preserves raw interactions, profile memory synthesizes high-level user attributes:

\begin{itemize}
    \item Demographic information volunteered by the user.
    \item Stated preferences and interests.
    \item Behavioral patterns observed across sessions.
    \item Professional context and domain expertise.
\end{itemize}

Profile memory is stored in SQL databases (PostgreSQL or SQLite) and supports both retrieval and update operations. When new information contradicts existing profile data, the system can update the profile to reflect the most recent state—supporting the \emph{knowledge update} capability evaluated in benchmarks such as LongMemEval~\cite{wu2024longmemeval}.

\subsection{Multi-Tenancy and Isolation}

MemMachine implements project-based namespace isolation. Each project is identified by an \texttt{org\_id/project\_id} pair and maintains separate memory instances. Sessions are further isolated by \texttt{user\_id}, \texttt{agent\_id}, and \texttt{session\_id}, enabling multi-tenant deployments where multiple users and agents operate with fully isolated memory stores.

\section{Retrieval Agent}
\label{sec:retrieval_agent}

MemMachine's baseline retrieval mechanism—vector similarity search with optional reranking—performs well on single-hop queries where the answer resides in a single episode cluster. However, production AI agents routinely encounter queries that require \emph{multi-hop reasoning}, \emph{multi-entity fan-out}, or \emph{cross-referential dependency chains}. For such queries, a single embedding cannot capture all the information needed because intermediate entities are unknown at query time. We call this the \textbf{late binding problem}: the correct query for a later retrieval hop cannot be formulated until an earlier hop has been resolved.

To address this, MemMachine v0.3 introduces the \textbf{Retrieval Agent}—an opt-in, LLM-orchestrated retrieval pipeline that routes queries to purpose-built strategies while maintaining bounded cost and latency. The Retrieval Agent augments (not replaces) the baseline search: callers who do not enable agent mode experience zero behavioral change.

\subsection{The Late Binding Problem}

Consider a multi-hop query: \emph{``What is the current employer of the spouse of the CEO of Acme?''} Answering this requires three ordered resolution steps: (1) identify the CEO of Acme $\to$ Person~X, (2) identify the spouse of Person~X $\to$ Person~Y, (3) identify the employer of Person~Y $\to$ Company~Z. At query time, only the original query string is available. Its embedding clusters around surface terms (``Acme,'' ``CEO,'' ``company'') and has no path to Company~Z because the intermediate entities are unknown. This is not a limitation of the embedding model—it is a structural property of information dependencies in multi-hop chains that no single-shot vector retrieval can resolve.

Existing mitigation strategies—query expansion (HyDE), BM25 hybrid search, chunk-level reranking—improve single-hop recall but cannot resolve dependency chains because they still operate on a single query formulation. Knowledge graph traversal solves late binding exactly but requires prior graph construction, which is expensive and lossy for arbitrary conversational content.

\subsection{Architecture}

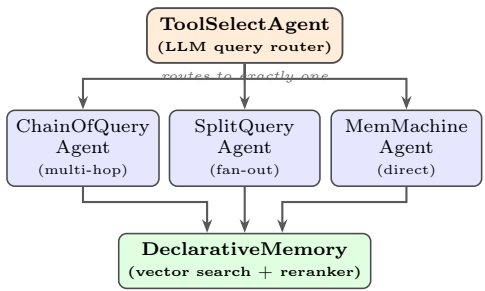
\begin{figure}[t]
\centering
\begin{tikzpicture}[
    every node/.style={font=\scriptsize},
    routerbox/.style={rectangle, draw=black!70, rounded corners=4pt, minimum width=2.6cm, minimum height=0.55cm, font=\scriptsize\bfseries, align=center, fill=orange!15, line width=0.7pt},
    stratbox/.style={rectangle, draw=black!70, rounded corners=3pt, minimum width=2.0cm, minimum height=0.7cm, font=\scriptsize, align=center, fill=blue!10, line width=0.6pt},
    leafbox/.style={rectangle, draw=black!70, rounded corners=3pt, minimum width=2.8cm, minimum height=0.55cm, font=\scriptsize\bfseries, align=center, fill=green!12, line width=0.6pt},
    arrow/.style={-{Stealth[length=1.8mm]}, thick, black!65},
]

\node[routerbox] (router) at (2.15, 0) {ToolSelectAgent\\{\tiny(LLM query router)}};

\node[stratbox] (chain) at (0, -1.5) {ChainOfQuery\\Agent\\{\tiny(multi-hop)}};
\node[stratbox] (split) at (2.15, -1.5) {SplitQuery\\Agent\\{\tiny(fan-out)}};
\node[stratbox] (direct) at (4.3, -1.5) {MemMachine\\Agent\\{\tiny(direct)}};

\node[leafbox] (decl) at (2.15, -3.0) {DeclarativeMemory\\{\tiny(vector search + reranker)}};

\draw[arrow] (router.south) -- ++(0,-0.2) -| (chain.north);
\draw[arrow] (router.south) -- (split.north);
\draw[arrow] (router.south) -- ++(0,-0.2) -| (direct.north);

\draw[arrow] (chain.south) -- ++(0,-0.2) -| ([xshift=-0.5cm]decl.north);
\draw[arrow] (split.south) -- (decl.north);
\draw[arrow] (direct.south) -- ++(0,-0.2) -| ([xshift=0.5cm]decl.north);

\node[font=\tiny\itshape, text=black!60] at (2.15, -0.55) {routes to exactly one};

\end{tikzpicture}
\caption{Retrieval Agent tool tree. The ToolSelectAgent classifies each query and routes it to one of three strategies. All strategies ultimately delegate to the same declarative memory search, ensuring that index and reranker improvements propagate automatically.}
\label{fig:retrieval_agent}
\end{figure}

The Retrieval Agent is implemented as a \textbf{composable tool tree} assembled inside MemMachine's long-term memory module. Figure~\ref{fig:retrieval_agent} shows the architecture: a root router dispatches each query to exactly one of three strategy nodes, all of which ultimately delegate to the same declarative memory search.

\textbf{Design principles.} All three strategy nodes delegate to the same \texttt{MemMachine} leaf, ensuring that improvements to the underlying vector search propagate automatically to all strategies. The tree is constructed once at startup and cached; routing decisions are made per-query at inference time. Agent mode is enabled via a single flag (\texttt{agent\_mode=true}), with zero behavioral change to existing callers. Each node exposes advisory cost properties (\texttt{accuracy\_score}, \texttt{token\_cost}, \texttt{time\_cost}) for future budget-aware routing.

\subsection{Query Routing}

The \textbf{ToolSelectAgent} (root node) uses a single LLM call to classify each incoming query into one of three structural types:

\begin{itemize}
    \item \textbf{Multi-hop dependency chain:} Two or more sequentially dependent retrieval steps where a later step requires the result of an earlier one. Routed to \texttt{ChainOfQuery}.
    \item \textbf{Single-hop multi-entity:} Multiple independent subjects answerable via parallel lookups with no inter-dependency. Routed to \texttt{SplitQuery}.
    \item \textbf{Single-hop direct:} Single subject, single lookup, no decomposition needed. Routed to \texttt{MemMachine} (baseline search with only the routing overhead).
\end{itemize}

The classification prompt uses embedded calibration examples and a tie-breaker rule: if any explicit dependency chain exists, classify as multi-hop even if multiple entities appear. This conservative bias trades extra LLM calls for higher recall completeness.

We further tuned Retrieval Agent prompts using the APO (Auto Prompt Optimization) algorithm from Agent Lightning~\cite{luo2025agentlightning}, using the Microsoft APO implementation\footnote{\url{https://microsoft.github.io/agent-lightning/latest/algorithm-zoo/apo/}}. In our internal ablations, tuning only the final answer prompt improved accuracy by approximately 4\% (with baseline prompt quality also improving), while jointly tuning all agent prompts improved accuracy by approximately 6\%. We do not perform live tuning at inference time, so these gains add no runtime token or latency overhead.

\subsection{Strategy Details}

\textbf{ChainOfQuery} implements iterative evidence accumulation for dependency chains. It executes up to 3 iterations, each consisting of: (1)~retrieval against the current query, (2)~a combined sufficiency judgment and query rewrite via a single LLM call, and (3)~evidence accumulation. The sufficiency prompt enforces evidence-only judgment (no external knowledge), strict completeness standards, entity grounding (rewrites use only entities present in retrieved evidence), and calibrated confidence scoring with early stopping at $\geq 0.8$ confidence. This design, grounded in the prompt engineering methodology of Luo et al.~\cite{luo2025agentlightning}, formalizes forward-chaining multi-hop reasoning with bounded cost.

\textbf{SplitQuery} addresses fan-out queries by decomposing them into 2--6 independent sub-queries via a single LLM call, executing all sub-queries concurrently via \texttt{asyncio.gather()}, and pooling the results. The decomposition prompt enforces structural constraints: sub-queries must each be answerable by a single fact lookup, derived operations (compare, rank, difference) are prohibited, and a conservative tie-breaker defaults to no-split when ambiguous.

\textbf{MemMachine} is the direct retrieval leaf that calls \texttt{DeclarativeMemory.search\_scored()} without any query transformation. It serves both as the leaf primitive for other agents' child calls and as the strategy for simple single-hop queries, where agent mode incurs only one routing LLM call overhead.

\subsection{Multi-Query Reranking}

A key innovation in agent mode is \textbf{multi-query reranking}: the final reranker receives a concatenation of \emph{all} queries used during retrieval (original query plus all rewrites or sub-queries), not just the original. This ensures that episodes relevant to any step in the retrieval chain—including intermediate facts that are critical for multi-hop reasoning but not directly referenced in the original query—score well in the final ranking.

\subsection{Benchmark Results}

We evaluate the Retrieval Agent across five benchmarks, comparing three modes: \textbf{Baseline} (LLM with full context, no memory), \textbf{MemMachine} (declarative memory search), and \textbf{Retrieval Agent} (agent-orchestrated search). Table~\ref{tab:agent_benchmarks} summarizes accuracy and recall across all benchmarks.

\begin{table*}[t]
\centering
\small
\caption{Retrieval Agent benchmark results across five datasets. Accuracy is LLM-judge score; Recall measures gold supporting fact retrieval. Best result per benchmark in \textbf{bold}. All MemMachine and Agent results use surrounding episodes enabled unless noted.}
\label{tab:agent_benchmarks}
\begin{tabular}{@{}llcccccc@{}}
\toprule
\textbf{Benchmark} & \textbf{Questions} & \textbf{Answer LLM} & \multicolumn{2}{c}{\textbf{MemMachine}} & \multicolumn{2}{c}{\textbf{Retrieval Agent}} & \textbf{Baseline} \\
\cmidrule(lr){4-5} \cmidrule(lr){6-7}
 & & & Acc. & Recall & Acc. & Recall & Acc. \\
\midrule
LoCoMo & 1,540 & gpt-5-mini & 90.5\% & 82.2\% & \textbf{90.2\%} & \textbf{82.5\%} & 91.7\% \\
WikiMultiHop & 500 & gpt-4.1-mini & 88.8\% & 90.8\% & \textbf{90.0\%} & \textbf{92.5\%} & 96.7\% \\
WikiMultiHop$^\dagger$ & 500 & gpt-5-mini & 87.4\% & 91.7\% & \textbf{92.6\%} & \textbf{83.3\%} & 96.7\% \\
MRCR & 300 & gpt-5.2 & 79.6\% & 99.1\% & \textbf{81.4\%} & \textbf{99.4\%} & 32.3\% \\
EpBench & 546 & gpt-5-mini & 73.4\% & 66.1\% & 71.8\% & 65.3\% & 77.5\% \\
EpBench & 546 & gpt-4o-mini & 71.4\% & 67.9\% & \textbf{73.3\%} & \textbf{69.7\%} & 50.1\% \\
HotpotQA & 500 & gpt-5-mini & 91.2\% & 91.0\% & \textbf{93.2\%} & \textbf{95.5\%} & 93.0\% \\
\bottomrule
\multicolumn{8}{@{}l@{}}{\scriptsize $^\dagger$WikiMultiHop fully randomized cross-question noise injection; all others with content in order.}
\end{tabular}
\end{table*}

\textbf{HotpotQA} provides the clearest demonstration of the Retrieval Agent's value. On the hard validation set (500 questions), the agent achieves 93.2\% overall accuracy and 92.31\% gold-supporting-fact recall, compared to 91.2\% accuracy and 90.98\% recall for baseline MemMachine—a +2.0 and +1.3 percentage point improvement respectively. The per-tool breakdown (Table~\ref{tab:agent_hotpot_tools}) shows that \texttt{ChainOfQuery} achieves the highest recall (95.31\%) on multi-hop bridge questions, validating the iterative evidence accumulation strategy.

\begin{table}[t]
\centering
\small
\caption{HotpotQA hard set: per-tool performance breakdown for the Retrieval Agent (n=500, gpt-5-mini).}
\label{tab:agent_hotpot_tools}
\begin{tabular}{@{}lrrr@{}}
\toprule
\textbf{Tool} & \textbf{Questions} & \textbf{Accuracy} & \textbf{Recall} \\
\midrule
MemMachine & 201 & 93.53\% & 89.31\% \\
SplitQuery & 118 & 94.07\% & 92.83\% \\
ChainOfQuery & 181 & 92.27\% & \textbf{95.31\%} \\
\midrule
Overall (Agent) & 500 & \textbf{93.20\%} & \textbf{92.31\%} \\
\bottomrule
\end{tabular}
\end{table}

\textbf{WikiMultiHop} demonstrates the benefit of agent orchestration under realistic noise conditions. When all question contexts are pooled into a single shared episodic store with fully randomized ordering—simulating a production setting where relevant and irrelevant memories are interleaved—the Retrieval Agent with gpt-5-mini achieves 92.6\% accuracy vs.\ 87.4\% for baseline MemMachine, a +5.2 point improvement.

\textbf{MRCR} (Multi-Round Co-reference Resolution) shows consistent agent improvement: 81.4\% vs.\ 79.6\% for MemMachine, with near-perfect recall (99.4\%). Notably, the LLM baseline without memory scores only 32.3\%, demonstrating that co-reference resolution fundamentally requires memory retrieval.

\textbf{LoCoMo} shows comparable performance between MemMachine (90.5\%) and the Retrieval Agent (90.2\%). This is expected: LoCoMo is predominantly single-hop conversational questions where baseline vector search is already effective, and the agent's routing overhead provides minimal benefit.

\textbf{EpBench} results are mixed and benchmark-dependent. With gpt-4o-mini as the answer model, the Retrieval Agent improves over baseline MemMachine (73.3\% vs.\ 71.4\%). With gpt-5-mini, baseline MemMachine slightly outperforms (73.4\% vs.\ 71.8\%), suggesting model-prompt sensitivity.

\subsection{Token Cost Analysis}

The Retrieval Agent's improved recall comes at the cost of additional LLM calls for routing and strategy execution. Table~\ref{tab:agent_tokens} shows the per-strategy token cost on HotpotQA.

\begin{table}[t]
\centering
\small
\caption{Average token cost per question by Retrieval Agent strategy (HotpotQA hard set).}
\label{tab:agent_tokens}
\begin{tabular}{@{}lrr@{}}
\toprule
\textbf{Component} & \textbf{Input Tokens} & \textbf{Output Tokens} \\
\midrule
ToolSelect (router) & 1,049 & 195 \\
ChainOfQuery & 2,874 & 1,614 \\
SplitQuery & 1,229 & 435 \\
MemMachine & 0 & 0 \\
\bottomrule
\end{tabular}
\end{table}

For the 36\% of queries routed directly to \texttt{MemMachine}, the total agent overhead is only the routing call ($\sim$1,244 tokens). For multi-hop queries routed to \texttt{ChainOfQuery}, total cost reaches $\sim$5,732 tokens per question—substantially more, but bounded by the 3-iteration limit. This bounded cost profile is a deliberate improvement over the predecessor unbounded agent loop (OpenAI Agents SDK with \texttt{max\_turns=30}), which was retired in favor of architecturally-defined retrieval strategies.

\subsection{When to Use Agent Mode}
\label{sec:when_agent}

Agent mode is not universally beneficial. Table~\ref{tab:agent_guidance} provides deployment guidance based on query characteristics.

\begin{table}[t]
\centering
\small
\caption{Decision guidance for enabling agent mode.}
\label{tab:agent_guidance}
\begin{tabular}{@{}p{2.35cm}p{3.25cm}@{}}
\toprule
\textbf{Use Agent Mode} & \textbf{When\ldots} \\
\midrule
\textbf{Best fit} & Queries involve multi-hop reasoning, co-reference resolution, or multi-entity fan-out. Accuracy is prioritized over latency. Use cases include research assistants, compliance review, and complex QA. \\
\addlinespace
\textbf{Not needed} & Queries are predominantly single-hop factual lookups. Latency requirements are strict ($<$200ms). Token budget is constrained. Use cases include simple chatbot recall, preference lookup, and session context. \\
\bottomrule
\end{tabular}
\end{table}

The system is designed for seamless coexistence: agent mode is enabled per-query via a single flag, and the router itself filters simple queries to the zero-overhead direct path. In production, applications can enable agent mode selectively based on query complexity heuristics or always-on with acceptable overhead for the majority of single-hop queries that route directly.

\subsection{OpenClaw Integration}

The Retrieval Agent is also available through \textbf{OpenClaw}\footnote{\url{https://github.com/openclaw}}, an open-source AI agent framework. The MemMachine OpenClaw integration is implemented as a standard plugin using the OpenClaw Plugin SDK (\texttt{openclaw/plugin-sdk}), not OpenClaw ``agent mode.'' Concretely, the plugin imports SDK types and exposes the canonical plugin object with \texttt{register(api: OpenClawPluginApi)}, where lifecycle hooks and tool registrations connect MemMachine capabilities to the host application. Through this plugin interface, applications can use both declarative memory search and Retrieval Agent orchestration via a unified tool surface. Benchmark results in Table~\ref{tab:agent_benchmarks} include OpenClaw-based evaluation configurations, demonstrating that the Retrieval Agent's benefits transfer across agent frameworks---not just MemMachine's native evaluation harness.

\section{LLM Integration and Model Impact}
\label{sec:llm_integration}

\subsection{How LLMs Are Used in MemMachine}

A distinguishing design principle of MemMachine is that LLMs are used sparingly and strategically, rather than for every memory operation. Specifically, LLMs serve three functions:

\begin{enumerate}
    \item \textbf{STM Summarization:} When the short-term memory window overflows, an LLM generates a compressed summary of the session context.
    \item \textbf{Profile Extraction:} An LLM extracts structured user facts and preferences from conversational data for profile memory.
    \item \textbf{Agent-Mode Inference:} When operating in agent mode, the eval-LLM can iteratively query MemMachine's memory as a tool to formulate responses.
\end{enumerate}

Critically, LLMs are \emph{not} used for per-message fact extraction, memory deduplication, or routine memory management—operations that in competing systems account for the majority of LLM token consumption.

\subsection{Impact of Model Choice on Performance}

The choice of LLM significantly affects both benchmark performance and operational cost. Table~\ref{tab:model_comparison} summarizes the impact of eval-LLM selection on LoCoMo scores.

\begin{table}[H]
\centering
\small
\caption{Impact of eval-LLM on MemMachine v0.2 LoCoMo scores.}
\label{tab:model_comparison}
\begin{tabular}{@{}lccc@{}}
\toprule
\textbf{Configuration} & \textbf{LLM Score} & \textbf{Mode} \\
\midrule
gpt-4.1-mini (agent) & 0.9169 & Agent \\
gpt-4.1-mini (memory) & 0.9123 & Memory \\
gpt-4o-mini (agent) & 0.8812 & Agent \\
gpt-4o-mini (memory) & 0.8747 & Memory \\
\bottomrule
\end{tabular}
\end{table}

The transition from gpt-4o-mini to gpt-4.1-mini yields a 3--4 percentage point improvement across both modes, demonstrating that newer LLM generations improve memory-augmented reasoning without any changes to the memory system itself. This also suggests that MemMachine's architecture is \emph{LLM-agnostic}—performance improvements in the underlying model translate directly to improved memory-augmented agent behavior.

\subsection{Cost Considerations}

Token usage is a primary cost driver for LLM-based applications. MemMachine's architecture substantially reduces token consumption:

\begin{table}[H]
\centering
\small
\caption{Token usage comparison on LoCoMo (gpt-4.1-mini).}
\label{tab:token_usage}
\begin{tabular}{@{}p{2.0cm}rr@{}}
\toprule
\textbf{System} & \textbf{Input Tokens} & \textbf{Output Tokens} \\
\midrule
MemMachine v0.2 (memory) & 4.20M & 43,169 \\
MemMachine v0.2 (agent) & 8.57M & 93,210 \\
Mem0 main/HEAD (memory) & 19.21M & 14,840 \\
\bottomrule
\end{tabular}
\end{table}

MemMachine uses approximately 78\% fewer input tokens than Mem0 in memory mode, translating directly to lower inference cost and reduced time-to-first-token latency.


\subsection{Context Window Considerations}

A natural question is whether memory systems provide benefit when conversational content fits entirely within the LLM's context window. Evidence from the LoCoMo benchmark suggests that even for conversations within context window limits (16K--26K tokens), memory-augmented systems outperform raw full-context baselines~\cite{maharana2024locomo}. This is because:

\begin{itemize}
    \item Full-context approaches suffer from the ``lost in the middle'' effect~\cite{liu2024lost}, where information in the middle of long contexts is poorly attended.
    \item Memory systems provide \emph{selective retrieval}, surfacing only the most relevant episodes rather than overwhelming the model with full conversation history.
    \item As conversations grow beyond context limits, memory systems become essential rather than merely beneficial.
\end{itemize}

\section{Experimental Setup}
\label{sec:setup}

\subsection{Benchmarks}

We evaluate MemMachine on the following benchmarks:

\textbf{LoCoMo}~\cite{maharana2024locomo}: Evaluates very long-term conversational memory across four question categories: single-hop (841 questions), multi-hop (282), temporal reasoning (321), and open-domain (96). Total: 1,540 scored questions (the 446 adversarial questions are excluded from scoring per standard practice). The evaluation code is based on Mem0's published evaluation framework.\footnote{\url{https://github.com/mem0ai/mem0/tree/main/evaluation}}


\textbf{LongMemEval\textsubscript{S}}~\cite{wu2024longmemeval}: Benchmarks five core long-term memory abilities---single-session information extraction (user-stated facts, assistant-stated facts, and preference inference), temporal reasoning, knowledge updates, and multi-session reasoning---using 500 curated questions embedded in chat histories of approximately 115k tokens each. We ingest the benchmark's chat histories session-by-session, then answer each question using MemMachine's memory search API. Evaluation uses the question-specific judge prompts provided by the LongMemEval framework. We conduct a systematic ablation study across 12 configurations varying six optimization dimensions.

\subsection{Evaluation Metrics}

For LoCoMo, we report three metrics:
\begin{itemize}
    \item \textbf{LLM Judge Score (\texttt{llm\_score}):} A binary score (0 or 1) assigned by a judge-LLM comparing the system's answer to ground truth. The overall score is the weighted mean across categories.
    \item \textbf{BLEU Score:} $n$-gram overlap with reference answers.
    \item \textbf{F1 Score:} Token-level precision and recall against reference answers.
\end{itemize}

The primary metric for comparison is the \texttt{llm\_score}, as it best captures semantic equivalence between generated and reference answers.

For LongMemEval\textsubscript{S}, we report the \texttt{llm\_score} per question category and overall, using the benchmark's standard LLM-judge evaluation with GPT-4o-mini as judge.

\subsection{System Environment}

\begin{table}[H]
\centering
\small
\caption{Benchmark test environment.}
\label{tab:environment}
\begin{tabular}{@{}p{2.35cm}p{3.15cm}@{}}
\toprule
\textbf{Component} & \textbf{Specification} \\
\midrule
Operating System & Ubuntu 24.04 LTS \\
CPU & 8 vCPUs \\
RAM & 16 GiB \\
GPU & Not required (CPU-only runs) \\
Python Version & 3.11 \\
MemMachine Version & v0.3.x \\
Database & PostgreSQL + Neo4j \\
Embedding Model & OpenAI text-embedding-3-small \\
Reranker & AWS Cohere rerank-v3-5:0 \\
Eval-LLM & OpenAI gpt-4o-mini / gpt-4.1-mini / gpt-5 / gpt-5-mini \\
Judge-LLM & OpenAI gpt-4o-mini \\
\bottomrule
\end{tabular}
\end{table}

\subsection{Compared Systems}

We compare MemMachine against the following systems using publicly reported results:

\begin{itemize}
    \item \textbf{Mem0}~\cite{chhikara2025mem0}: Tested with Mem0 main/HEAD and re-run with gpt-4.1-mini for fair comparison.
    \item \textbf{Zep}~\cite{rasmussen2025zep}: Results sourced from Zep's published evaluation.
    \item \textbf{Memobase}: Results sourced from Memobase's published benchmark.
    \item \textbf{LangMem}: Results from Mem0's comparative evaluation.
    \item \textbf{OpenAI baseline}: ChatGPT's native memory.
\end{itemize}

\subsection{Reproducibility}

All benchmark scripts, configuration files, and run instructions are available in the MemMachine repository under \texttt{evaluation/}.\footnote{\url{https://github.com/MemMachine/MemMachine/tree/main/evaluation}} For reproducible reporting, we recommend pinning a repository tag or commit hash, recording model versions and API provider settings, and saving raw per-question outputs used for score aggregation. Researchers can reproduce our results with their own hardware and API keys.

\section{Results and Analysis}
\label{sec:results}

\subsection{LoCoMo Benchmark Results}

Table~\ref{tab:locomo_results} presents the comprehensive LoCoMo results for MemMachine v0.2 across both LLM configurations and both operating modes.

\begin{table*}[t]
\centering
\small
\caption{MemMachine v0.2 LoCoMo results by category (LLM Judge Score).}
\label{tab:locomo_results}
\begin{tabular}{@{}llccccc@{}}
\toprule
\textbf{Eval-LLM} & \textbf{Mode} & \textbf{Multi-hop} & \textbf{Temporal} & \textbf{Open-domain} & \textbf{Single-hop} & \textbf{Overall} \\
\midrule
gpt-4.1-mini & Agent & 0.8830 & 0.9159 & 0.7188 & 0.9512 & \textbf{0.9169} \\
gpt-4.1-mini & Memory & 0.8972 & 0.8910 & 0.7500 & 0.9441 & 0.9123 \\
gpt-4o-mini & Agent & 0.8404 & 0.8069 & 0.7396 & 0.9394 & 0.8812 \\
gpt-4o-mini & Memory & 0.8759 & 0.7352 & 0.7083 & 0.9465 & 0.8747 \\
\bottomrule
\end{tabular}
\end{table*}

\subsection{Comparative Analysis}

Table~\ref{tab:comparison} compares MemMachine against competing systems on LoCoMo.

\begin{table*}[t]
\centering
\small
\caption{LoCoMo benchmark comparison across AI agent memory systems (LLM Judge Score). MemMachine results are with gpt-4o-mini for fair comparison with published baselines; gpt-4.1-mini results shown separately in Table~\ref{tab:locomo_results}.}
\label{tab:comparison}
\begin{tabular}{@{}lccccc@{}}
\toprule
\textbf{System} & \textbf{Single-hop} & \textbf{Temporal} & \textbf{Multi-hop} & \textbf{Open-domain} & \textbf{Overall} \\
\midrule
\textbf{MemMachine v0.2} & \textbf{0.9465} & 0.7352 & \textbf{0.8759} & 0.7083 & \textbf{0.8747} \\
Memobase (v0.0.37) & 0.7092 & \textbf{0.8505} & 0.4688 & \textbf{0.7717} & 0.7578 \\
Zep & 0.7411 & 0.7979 & 0.6604 & 0.6771 & 0.7514 \\
Mem0 & 0.6713 & 0.5551 & 0.5115 & 0.7293 & 0.6688 \\
LangMem & 0.6223 & 0.2343 & 0.4792 & 0.7112 & 0.5810 \\
OpenAI & 0.6379 & 0.2171 & 0.4292 & 0.6229 & 0.5290 \\
\bottomrule
\end{tabular}
\end{table*}

In our LoCoMo comparison setting, MemMachine achieves the highest overall score by +9.7 points over the next-best system (Memobase). Key observations:

\begin{itemize}
    \item \textbf{Single-hop (0.9465):} MemMachine's sentence-level indexing and ground truth preservation enable exceptional factual recall.
    \item \textbf{Multi-hop (0.8759):} The contextualization mechanism allows linking related information across sessions.
    \item \textbf{Temporal (0.7352):} Competitive but trailing Memobase (0.8505), suggesting room for improvement in temporal reasoning—likely addressable through enhanced timestamp-aware retrieval.
    \item \textbf{Open-domain (0.7083):} Strong performance considering that episodic memory is optimized for user-centric rather than world-knowledge queries.
\end{itemize}

With gpt-4.1-mini, MemMachine's temporal score improves to 0.9159 in agent mode, suggesting that eval-model capability is a major factor in temporal reasoning outcomes.

\subsection{Efficiency Analysis}

Beyond accuracy, MemMachine demonstrates substantial efficiency advantages:

\begin{itemize}
    \item \textbf{Token Reduction:} $\sim$80\% fewer input tokens than Mem0, directly reducing API costs.
    \item \textbf{Memory Add Speed:} $\sim$75\% faster than previous versions, enabling real-time ingestion.
    \item \textbf{Search Speed:} Up to 75\% faster search operations, reducing end-to-end response latency.
\end{itemize}

\subsection{\texorpdfstring{LongMemEval\textsubscript{S}}{LongMemEval-S} Ablation Study}
\label{sec:longmemeval}

We evaluate MemMachine on the full LongMemEval\textsubscript{S} benchmark ($n=500$ questions) through a systematic ablation study across six optimization dimensions: sentence chunking, user-query bias correction, context formatting, retrieval depth ($k$), search prompt design, and answer-model selection. Each dimension is evaluated by comparing configuration pairs that differ in exactly one variable. Table~\ref{tab:lme_configs} summarizes the key configurations; Table~\ref{tab:lme_ablation} isolates the contribution of each optimization.

\begin{table*}[t]
\centering
\small
\caption{LongMemEval\textsubscript{S} configuration matrix ($n=500$ questions per run). ``Chunk'' = sentence-level chunking enabled; ``User-Q'' = user-role query prefix; ``JSON-str'' = structured message formatting; ``Edwin\{1,3\}'' = search prompt variants of increasing refinement.}
\label{tab:lme_configs}
\begin{tabular}{@{}llcccccr@{}}
\toprule
\textbf{ID} & \textbf{Answer LLM} & \textbf{Chunk} & \textbf{User-Q} & \textbf{JSON-str} & \textbf{Prompt} & \textbf{top\_$k$} & \textbf{LLM Score} \\
\midrule
C5  & GPT-5       & ---   & ---     & \checkmark & Edwin1 & 20  & 0.855 \\
C6  & GPT-5       & ---   & \checkmark & \checkmark & Edwin1 & 20  & 0.870 \\
C7  & GPT-5       & ---   & \checkmark & \checkmark & Edwin1 & 30  & 0.912 \\
C8  & GPT-5       & ---   & \checkmark & \checkmark & Edwin1 & 50  & 0.890 \\
C9  & GPT-5       & \checkmark & \checkmark & \checkmark & Edwin1 & 20  & 0.878 \\
C11 & GPT-5       & \checkmark & \checkmark & \checkmark & Edwin3 & 20  & 0.896 \\
C12 & GPT-5-mini  & \checkmark & \checkmark & \checkmark & Edwin3 & 20  & 0.922 \\
C13 & GPT-5-mini  & \checkmark & \checkmark & \checkmark & Edwin3 & 30  & 0.916 \\
C14 & GPT-5-mini  & \checkmark & \checkmark & \checkmark & Edwin3 & 50  & 0.928 \\
C15 & GPT-5-mini  & \checkmark & \checkmark & \checkmark & Edwin3 & 100 & \textbf{0.930} \\
C16 & GPT-5       & \checkmark & \checkmark & \checkmark & Edwin3 & 30  & 0.902 \\
C17 & GPT-5       & \checkmark & \checkmark & \checkmark & Edwin3 & 50  & 0.914 \\
\bottomrule
\end{tabular}
\end{table*}

\begin{table}[t]
\centering
\small
\caption{LongMemEval\textsubscript{S} ablation: contribution of each optimization to overall LLM score, measured by comparing configuration pairs that differ in one variable.}
\label{tab:lme_ablation}
\begin{tabular}{@{}lcc@{}}
\toprule
\textbf{Optimization} & \textbf{Comparison} & \textbf{$\Delta$ Score} \\
\midrule
\multicolumn{3}{@{}l}{\emph{Retrieval-stage optimizations}} \\
\addlinespace[2pt]
Retrieval depth ($k$: 20$\to$30)     & C6 vs.\ C7 & +4.2\% \\
Context formatting (JSON-str)        & C4 vs.\ C5 & +2.0\% \\
Search prompt (Edwin1$\to$3)         & C9 vs.\ C11 & +1.8\% \\
COT $\to$ simple prompt (GPT-5)      & C1 vs.\ C4 & +1.6\% \\
User-query bias correction           & C5 vs.\ C6 & +1.4\% \\
\addlinespace[3pt]
\multicolumn{3}{@{}l}{\emph{Ingestion-stage optimization}} \\
\addlinespace[2pt]
Sentence chunking                    & C6 vs.\ C9 & +0.8\% \\
\addlinespace[3pt]
\multicolumn{3}{@{}l}{\emph{Model selection}} \\
\addlinespace[2pt]
GPT-5 $\to$ GPT-5-mini              & C11 vs.\ C12 & +2.6\% \\
\bottomrule
\end{tabular}
\end{table}

\subsubsection{\texorpdfstring{Retrieval Depth ($k$)}{Retrieval Depth (k)}}

The most impactful single parameter is the number of retrieved episodes. Increasing $k$ from 20 to 30 yields the largest improvement: +4.2 percentage points (C6: 0.870~$\to$~C7: 0.912). However, further increases show diminishing or negative returns: $k=50$ drops to 0.890, \emph{worse} than $k=30$.

This non-monotonic behavior reflects a tension between recall and noise. At $k=20$, the retrieval window misses some relevant episodes, particularly for multi-hop questions requiring evidence from multiple sessions. At $k=30$, sufficient evidence is captured without overwhelming the answer LLM with distractors. At $k=50$, additional episodes introduce irrelevant context that degrades reading comprehension, consistent with the ``lost in the middle'' phenomenon~\cite{liu2024lost}.

Notably, this $k$-sensitivity is model-dependent. With GPT-5-mini (C12--C15), performance improves monotonically from $k=20$ (0.922) through $k=50$ (0.928) to $k=100$ (0.930), though with diminishing marginal gains. This suggests GPT-5-mini is more robust to distractor context than GPT-5, possibly due to differences in attention mechanisms or instruction-following behavior at high context lengths.

\subsubsection{User-Query Bias Correction}

We observe that MemMachine's search results can exhibit a bias toward retrieving assistant messages over user messages. Assistant messages are typically longer with more sentences and therefore more embedding keys, while user messages are shorter but often contain first-hand factual statements with higher informational value for recall tasks. Prepending the prefix \texttt{"user:"} to the search query shifts retrieval toward user messages, yielding +1.4\% (C5: 0.855~$\to$~C6: 0.870).

\subsubsection{Context Formatting}

The format in which retrieved messages are presented to the answer LLM significantly affects comprehension. A naive approach concatenating all messages with carriage returns produces a wall of text. Using \texttt{\textbackslash n} as line terminators \emph{within} a message while using actual carriage returns to \emph{separate} messages improves the LLM's ability to parse message boundaries, yielding +2.0\%.

\subsubsection{Sentence Chunking}

MemMachine's sentence-level chunking creates one embedding key per sentence rather than per message, producing finer-grained index entries. Enabling chunking yields +0.8\% (C6: 0.870~$\to$~C9: 0.878), a modest but consistent gain. The relatively small effect suggests that sentence-level granularity primarily helps on questions where the relevant information is a single sentence embedded within a longer message.

\subsubsection{Search Prompt Design}

We evaluate three search prompt variants (Edwin1, Edwin2, Edwin3) with increasing refinement. Edwin3 yields +1.8\% over Edwin1 (C9: 0.878~$\to$~C11: 0.896), demonstrating that the framing of the search query---not just the retrieval algorithm---materially affects recall quality.

\subsubsection{Answer LLM Selection}

A surprising finding is that GPT-5-mini outperforms GPT-5 as the answer LLM by +2.6\% (C11: 0.896~$\to$~C12: 0.922). The advantage persists across retrieval depths: at $k=30$, GPT-5-mini achieves 0.916 vs.\ GPT-5's 0.902; at $k=50$, 0.928 vs.\ 0.914. We attribute this to the interaction between prompt design and model architecture. The Edwin3 prompt is a direct, concise instruction without chain-of-thought scaffolding, which aligns with GPT-5-mini's streamlined instruction-following. Conversely, GPT-5's built-in reasoning may interfere when given explicit reasoning instructions. Since GPT-5-mini is also substantially cheaper per token, the best configuration is the most cost-efficient.

\subsubsection{Per-Category Analysis}

Table~\ref{tab:lme_percategory} presents per-category scores for selected configurations.

\begin{table*}[t]
\centering
\small
\caption{LongMemEval\textsubscript{S} per-category LLM scores for selected configurations ($n=500$). SSU = single-session-user, SSP = single-session-preference, SSA = single-session-assistant, TR = temporal reasoning, KU = knowledge update, MS = multi-session.}
\label{tab:lme_percategory}
\begin{tabular}{@{}llccccccc@{}}
\toprule
\textbf{Config} & \textbf{LLM} & \textbf{SSU} & \textbf{SSP} & \textbf{SSA} & \textbf{TR} & \textbf{KU} & \textbf{MS} & \textbf{Total} \\
\midrule
C5 (baseline)     & GPT-5      & 0.986 & 0.700 & 1.000 & 0.798 & 0.925 & 0.797 & 0.855 \\
C6 (user-q)       & GPT-5      & 0.957 & 0.667 & 0.982 & 0.850 & 0.885 & 0.835 & 0.870 \\
C7 ($k$=30)       & GPT-5      & 0.986 & 0.800 & 1.000 & 0.917 & 0.923 & 0.850 & 0.912 \\
C12 (mini,$k$=20) & GPT-5-mini & 0.986 & 0.933 & 1.000 & 0.902 & 0.962 & 0.850 & 0.922 \\
C14 (mini,$k$=50) & GPT-5-mini & 1.000 & 0.933 & 1.000 & 0.932 & 0.949 & 0.842 & 0.928 \\
C15 (mini,$k$=100)& GPT-5-mini & 1.000 & 0.933 & 0.982 & 0.917 & 0.949 & 0.872 & \textbf{0.930} \\
\bottomrule
\end{tabular}
\end{table*}

Several patterns emerge. \textbf{Single-session extraction} (SSU, SSA) is nearly saturated: most configurations achieve 0.98--1.00, indicating that sentence-level indexing is well-suited for recalling specific facts from individual sessions.
\textbf{Single-session preference} (SSP) shows the most dramatic improvement, rising from 0.700 (C5) to 0.933 (C12--C15). Preference questions require inferring user preferences from indirect cues, which benefits from both better retrieval and better answer models.
\textbf{Temporal reasoning} (TR) improves steadily with retrieval depth and prompt optimization, from 0.798 (C5) to 0.932 (C14), as retrieving more context helps establish temporal relationships.
\textbf{Multi-session reasoning} (MS) remains the most challenging category, peaking at 0.872 (C15, $k$=100), since synthesizing information across sessions requires the broadest retrieval window.

\subsubsection{Token Cost--Accuracy Tradeoff}

Retrieval depth directly affects token consumption. Table~\ref{tab:lme_tokens} shows the cost profile for selected configurations.

\begin{table}[t]
\centering
\small
\caption{LongMemEval\textsubscript{S} token consumption per 500-question run. Input tokens scale with $k$; output tokens remain approximately constant.}
\label{tab:lme_tokens}
\begin{tabular}{@{}lcrr@{}}
\toprule
\textbf{Config (LLM)} & \textbf{$k$} & \textbf{Input (M)} & \textbf{Score} \\
\midrule
C6 (GPT-5)       & 20  & 2.94  & 0.870 \\
C7 (GPT-5)       & 30  & 4.03  & 0.912 \\
C8 (GPT-5)       & 50  & 6.47  & 0.890 \\
C12 (GPT-5-mini) & 20  & 2.58  & 0.922 \\
C14 (GPT-5-mini) & 50  & 5.97  & 0.928 \\
C15 (GPT-5-mini) & 100 & 9.79  & 0.930 \\
\bottomrule
\end{tabular}
\end{table}

The Pareto-optimal configuration is C12 (GPT-5-mini, $k$=20): it achieves 0.922 with only 2.58M input tokens, outperforming C7 (GPT-5, $k$=30, 0.912) which requires 4.03M tokens. Reaching the peak score of 0.930 (C15) requires $3.8\times$ the input tokens of C12 for only +0.8\% accuracy.

\section{Discussion}
\label{sec:discussion}

\subsection{Retrieval Stage Dominates Accuracy}

Our LongMemEval ablation reveals that retrieval-stage optimizations contribute substantially more to final accuracy than ingestion-stage changes. The cumulative effect of retrieval-side improvements (retrieval depth: +4.2\%, formatting: +2.0\%, search prompt: +1.8\%, COT removal: +1.6\%, user-query bias: +1.4\%) far exceeds the ingestion-side contribution of sentence chunking (+0.8\%). This suggests that for memory systems, \emph{how} data is recalled matters more than \emph{how} it is stored, provided the storage preserves ground truth.

This has architectural implications: systems that invest heavily in LLM-based ingestion (extracting facts, building knowledge graphs) may be over-optimizing the wrong stage. MemMachine's approach of storing raw data with lightweight indexing and investing in retrieval quality appears to be more effective per unit of engineering effort.

\subsection{Model--Prompt Co-optimization}

The GPT-5-mini result on LongMemEval highlights that model selection and prompt design must be co-optimized. A chain-of-thought prompt designed for GPT-4.x is suboptimal for GPT-5, and a simple direct prompt can outperform a complex one when paired with the right model. The advantage persists across retrieval depths (GPT-5-mini beats GPT-5 by +1.4\% at $k$=30 and $k$=50). This argues against the common practice of reusing prompts across model upgrades, and suggests that memory system deployments should re-evaluate prompts whenever the underlying answer model changes.

\subsection{The Role of Personalization}

Personalization is perhaps the most compelling reason for AI agent memory. Without memory, every interaction starts from zero—the agent has no awareness of the user's history, preferences, or context. Memory transforms a generic LLM into a personalized assistant that adapts to individual users over time.

MemMachine's dual memory architecture directly supports personalization: episodic memory provides the factual grounding for ``what happened,'' while profile memory captures the distilled ``who the user is.'' This combination enables agents to:

\begin{itemize}
    \item Maintain continuity across sessions without requiring users to repeat context.
    \item Adapt responses to user preferences, communication style, and domain expertise.
    \item Build trust through demonstrated recall of past interactions.
    \item Provide proactive suggestions based on accumulated user knowledge.
\end{itemize}

As AI agents move from novelty to daily utility, personalization through memory will become a differentiating capability. Users will expect their agents to ``know'' them—and memory systems like MemMachine provide the infrastructure to deliver this expectation.

\subsection{Summary vs.\ Full Context vs.\ Compressed Observations}

A recurring design question is whether to provide the LLM with a summary of past interactions, the full conversational context, or a compressed intermediate form. Our findings, together with recent work on observational memory~\cite{mastra2026observational}, suggest that each approach occupies a distinct point in the design space:

\begin{itemize}
    \item \textbf{Full context} overwhelms the model, triggers the ``lost in the middle'' effect~\cite{liu2024lost}, and becomes infeasible as history grows beyond context limits.
    \item \textbf{Summary-only} approaches (compaction) lose critical detail, particularly for factual and temporal queries where the exact wording or timing matters. As Mastra's research notes, compaction produces ``documentation-style summaries'' that ``strip out the specific decisions and tool interactions agents need.''
    \item \textbf{Compressed observations} (Mastra) offer a middle ground: event-based logs that preserve structure while achieving 3--40$\times$ compression. This enables prompt caching and stable context windows but sacrifices the ability to retrieve original episodes on demand.
    \item \textbf{MemMachine's approach}—STM summary \emph{plus} selectively retrieved raw episodes—provides a different tradeoff: the summary gives high-level context, while retrieved episodes supply \emph{uncompressed} factual grounding. This is particularly important for use cases requiring auditability, compliance, or multi-hop reasoning over exact conversational records.
\end{itemize}

The choice between these approaches depends on the deployment context. For tool-heavy agents generating large outputs (coding agents, SRE agents), compression-first approaches may be optimal. For agents serving domains where factual precision matters (healthcare, legal, financial services), ground-truth-preserving retrieval is essential. A promising future direction is hybrid architectures that combine compressed observations for high-level context with on-demand retrieval of raw episodes when precision is needed.

\subsection{Single-Agent vs.\ Multi-Agent Memory}

Memory benefits increase in multi-agent environments:

\begin{itemize}
    \item \textbf{Shared memory} enables agents to coordinate without redundant information gathering.
    \item \textbf{Specialized agents} can deposit domain-specific knowledge that other agents retrieve, enabling emergent division of labor.
    \item \textbf{Session continuity} allows agent handoffs without context loss.
    \item \textbf{Reduced token usage} across the system, as agents share rather than regenerate context.
\end{itemize}

MemMachine's multi-tenancy architecture (project/session isolation) naturally supports multi-agent deployments where agents share a project-level memory while maintaining session-level isolation for individual conversations.

\subsection{Privacy and Data Sovereignty}

Memory systems for AI agents raise important privacy considerations. When user conversations are stored and indexed, the data sovereignty question becomes critical:

\begin{itemize}
    \item \textbf{Local LLMs:} Running embedding models and LLMs locally (e.g., using Ollama or vLLM) keeps all data on-premises. MemMachine supports local providers through its configurable model architecture.
    \item \textbf{Hosted APIs:} Using OpenAI, Google, or AWS services means conversational data traverses third-party infrastructure, subject to their data processing agreements.
    \item \textbf{Hybrid approaches:} Memory storage can remain local while only anonymized or aggregated queries are sent to hosted LLMs for summarization or inference.
\end{itemize}

MemMachine's open-source, self-hosted architecture gives organizations full control over their data pipeline. The configurable provider system allows swapping between local and hosted models without code changes, enabling organizations to match their privacy requirements to their deployment model.

\subsection{Limitations and Threats to Validity}

Our results should be interpreted with several limitations in mind. First, benchmark outcomes are sensitive to eval-model choice, prompt templates, and provider-side model updates; scores reported here are tied to the specific configurations listed in Section~\ref{sec:setup}. Second, cross-system comparisons mix re-run results and published numbers, which may differ in preprocessing, prompt settings, or infrastructure. Third, while LoCoMo, LongMemEval\textsubscript{S}, HotpotQA, WikiMultiHop, and EpBench cover important retrieval behaviors, they do not fully represent all production workloads (for example, multilingual, multimodal, or strict real-time constraints). Fourth, token-efficiency comparisons are workload-dependent and should be treated as directional outside the reported setup. Fifth, the LongMemEval ablation treats each optimization dimension independently; interaction effects between dimensions (e.g., whether chunking benefits change at higher $k$) remain unexplored, and the ablation configurations C1--C4 used partial question subsets before being extended to the full 500-question evaluation. We therefore view these results as strong empirical evidence within the evaluated settings rather than universal performance guarantees.

\subsection{Architectural Design Tensions}

The VentureBeat analysis of emerging memory architectures~\cite{mastra2026observational} identifies several key questions that enterprises should consider when selecting a memory approach. We frame these as design tensions that inform MemMachine's positioning:

\textbf{Retrieval vs.\ stable context.} Retrieval-based systems (MemMachine, Mem0, Zep) search for relevant memories each turn, which enables access to arbitrarily large memory stores but invalidates prompt caches and adds latency. Stable-context systems (Mastra's observational memory) keep a compressed log always in context, enabling prompt caching but capping the total memory to what fits in the context window. MemMachine's approach provides the scalability of retrieval while its STM component ensures that recent context is always immediately available without a retrieval step.

\textbf{Prompt cacheability.} Modern LLM providers (OpenAI, Anthropic) offer significant discounts for cached prompt prefixes. Systems that maintain a stable prefix can exploit this for 50--90\% cost reduction on cached tokens. MemMachine's STM summary provides a semi-stable prefix, though retrieved episodes vary per query. This represents an area for future optimization—for example, caching frequently retrieved episode clusters.

\textbf{Infrastructure complexity.} MemMachine requires database infrastructure (PostgreSQL, Neo4j) but provides full control over data persistence and query patterns. Text-only approaches (Mastra) eliminate the need for specialized databases but may become constrained as memory volume grows. For enterprises with existing database infrastructure, MemMachine's approach integrates naturally; for teams seeking minimal infrastructure, simpler architectures may be preferable initially.

\textbf{Memory as a top-level primitive.} There is growing consensus that memory is one of the essential primitives for production AI agents, alongside tool use, workflow orchestration, observability, and guardrails. MemMachine's design as a standalone, framework-agnostic memory layer—accessible via REST API, Python SDK, and MCP—reflects this view. Agents should be able to adopt memory without being locked into a specific orchestration framework.

Table~\ref{tab:design_space} summarizes the architectural tradeoffs across representative memory systems.

\begin{table*}[t]
\centering
\small
\caption{Architectural design space comparison across agent memory systems.}
\label{tab:design_space}
\begin{tabular}{@{}lcccccc@{}}
\toprule
\textbf{Property} & \textbf{MemMachine} & \textbf{Mem0} & \textbf{Zep} & \textbf{Mastra OM} & \textbf{MemOS} & \textbf{Full Context} \\
\midrule
Memory approach & Retrieval & Retrieval & Retrieval & In-context & Hybrid & In-context \\
Ground truth preserved & \checkmark & Partial & Partial & $\times$ & Partial & \checkmark \\
Prompt cacheable & Partial & $\times$ & $\times$ & \checkmark & $\times$ & \checkmark \\
Scales beyond context window & \checkmark & \checkmark & \checkmark & $\times$ & \checkmark & $\times$ \\
LLM calls per message & Low & High & Moderate & Moderate & High & None \\
Specialized DB required & Yes & Yes & Yes & No & Yes & No \\
Parametric/KV-cache memory & $\times$ & $\times$ & $\times$ & $\times$ & \checkmark & $\times$ \\
Works with closed-source LLMs & \checkmark & \checkmark & \checkmark & \checkmark & Partial & \checkmark \\
Open source & \checkmark & Partial & Partial & \checkmark & \checkmark & N/A \\
\bottomrule
\end{tabular}
\end{table*}

\subsection{When Memory Helps (and When It Doesn't)}

Memory provides clear benefits for:

\begin{itemize}
    \item Multi-session interactions (customer support, healthcare, education).
    \item Personalization-dependent applications (content recommendations, personal assistants).
    \item Complex workflows requiring state persistence (project management, CRM).
    \item Compliance and audit scenarios requiring interaction history.
\end{itemize}

Memory may be unnecessary or counterproductive for:

\begin{itemize}
    \item Single-turn, stateless queries (search, translation, simple QA).
    \item High-volume, low-personalization tasks (batch processing, data extraction).
    \item Scenarios where privacy constraints prohibit storing interaction history.
\end{itemize}

\section{Future Work}
\label{sec:future}

Several directions merit investigation:

\begin{itemize}
    \item \textbf{Procedural memory:} Extending MemMachine to store and retrieve learned action patterns, tool-use strategies, and workflow recipes.
    \item \textbf{Enhanced temporal reasoning:} Developing dedicated temporal indexing and query expansion techniques to improve performance on temporal benchmarks.
    \item \textbf{LongMemEval\textsubscript{M} evaluation:} Extending evaluation to LongMemEval\textsubscript{M} (500 sessions, $\sim$1.5M tokens per question), which tests memory at production scale.
    \item \textbf{Adaptive retrieval depth:} Implementing query-complexity-aware $k$ selection, informed by our finding that optimal $k$ depends on both the query type and the downstream answer model.
    \item \textbf{Memory consolidation and forgetting:} Implementing cognitive-inspired mechanisms for prioritizing frequently accessed memories and gracefully retiring stale information.
    \item \textbf{Multi-modal memory:} Supporting images, audio, and structured data alongside conversational text.
    \item \textbf{Additional database backends:} Expanding support to ChromaDB, Milvus, and other vector stores.
    \item \textbf{Reinforcement learning integration:} Using benchmark feedback to optimize retrieval strategies through learned policies.
    \item \textbf{Retrieval Agent extensions:} Expanding the agent tool tree with specialized agents for temporal reasoning, aggregation queries, and comparative analysis. Implementing budget enforcement (token cost ceilings, latency limits) with automatic fallback to cheaper strategies. Enabling per-agent LLM tier selection for cost optimization.
    \item \textbf{Adaptive retrieval budgets:} Dynamically adjusting per-sub-query retrieval limits based on query complexity estimates and accumulated evidence, reducing redundant episode retrieval in fan-out and chain-of-query strategies.
    \item \textbf{Function-calling code mode:} Investigating function-calling architectures where agents emit structured executable code (for example, Python or TypeScript) rather than invoking large predefined tool lists. Code executes in a secure interpreter that handles data processing, dynamic tool discovery, and multi-step chaining with fewer repeated LLM roundtrips. Prior reports indicate large token savings for massive toolsets (for example, 98.7\% in Anthropic's MCP code execution workflow and up to 99.9\% in Cloudflare's Code Mode)~\cite{anthropic2025mcp_codeexec,cloudflare2025codemode}. We will evaluate both client-side and server-side variants, including dynamic directory-based schema loading and low-overhead search/execute proxy patterns.
\end{itemize}

\section{Conclusion}
\label{sec:conclusion}

We have presented MemMachine, an open-source memory system for AI agents that prioritizes ground truth preservation, cost efficiency, and personalization. Through a two-tier architecture of short-term and long-term episodic memory augmented by profile memory, MemMachine provides agents with the ability to store, recall, and reason over past experiences without the high cost and error accumulation inherent in LLM-dependent extraction approaches.

Our evaluation spans multiple benchmarks. On LoCoMo, MemMachine achieves state-of-the-art performance (0.9169 with gpt-4.1-mini) with approximately 80\% fewer tokens than Mem0. On LongMemEval\textsubscript{S}, a systematic ablation across six optimization dimensions achieves 93.0\% overall accuracy and reveals that retrieval-stage optimizations---particularly retrieval depth tuning (+4.2\%) and context formatting (+2.0\%)---dominate over ingestion-stage changes, and that smaller models (GPT-5-mini) outperform larger models (GPT-5) when co-optimized with appropriate prompts. The Retrieval Agent extends these capabilities to multi-hop queries, achieving 93.2\% on HotpotQA hard and 92.6\% on WikiMultiHop with randomized noise---demonstrating that MemMachine's ground-truth-preserving architecture is composable: intelligent retrieval strategies can be layered on top without modifying the underlying storage model.

As AI agents transition from experimental technology to production infrastructure, the quality of their memory systems will determine the quality of their personalization, accuracy, and trustworthiness. MemMachine provides a foundation for this next generation of memory-augmented agents.

\bibliographystyle{unsrtnat}

\end{document}